\pdfoutput=1

\documentclass[11pt]{article}

\usepackage{authblk}
\usepackage{acl}

\usepackage{times}
\usepackage{latexsym}
\usepackage[T1]{fontenc}
\usepackage[utf8]{inputenc}
\usepackage{microtype}
\usepackage{graphicx}
\usepackage{adjustbox}
\usepackage{subfig}
\usepackage[super]{nth}
\usepackage{float}
\usepackage{mathtools}
\usepackage{amsmath}
\usepackage{inconsolata}
\usepackage{cleveref}
\usepackage{array}
\usepackage{pifont}
\usepackage{tabularx}
\usepackage{multirow}
\usepackage{enumitem}
\usepackage{xspace}
\usepackage{tcolorbox}
\usepackage{booktabs,amsfonts,dcolumn}
\usepackage{hyperref}
\usepackage{url}
\usepackage{amsmath,amsthm,amsfonts,amssymb,bm,stmaryrd,bbm}
\usepackage[T1]{fontenc}
\newcommand{\squishlist}{
   \begin{list}{$\bullet$}
    { \setlength{\itemsep}{0pt}      \setlength{\parsep}{3pt}
      \setlength{\topsep}{3pt}       \setlength{\partopsep}{0pt}
      \setlength{\leftmargin}{1.5em} \setlength{\labelwidth}{1em}
      \setlength{\labelsep}{0.5em} } }

\newcommand{\squishlisttwo}{
   \begin{list}{$\bullet$}
    { \setlength{\itemsep}{0pt}    \setlength{\parsep}{0pt}
      \setlength{\topsep}{0pt}     \setlength{\partopsep}{0pt}
      \setlength{\leftmargin}{2em} \setlength{\labelwidth}{1.5em}
      \setlength{\labelsep}{0.5em} } }

\newcommand{\squishend}{
    \end{list}  }
\usepackage{etoolbox}
\newcommand\blfootnote[1]{%
  \begingroup
  \renewcommand\thefootnote{}\footnote{#1}%
  \addtocounter{footnote}{-1}%
  \endgroup
}

\newsavebox\affbox
\usepackage{xspace}
\def\ie{i.e.}

\begin{document}
\title{Tell2Design: A Dataset for Language-Guided Floor Plan Generation}

%\author[1]{\parbox{2cm}{Author A \\ Author B}}
% \author[1,*]{Sicong Leng}
% \author[2,*,†]{Yang Zhou}
% \author[3]{Mohammed Haroon Dupty}
% \author[3]{Wee Sun Lee}
% \author[4]{Sam Conrad Joyce}
% \author[1]{Wei Lu}

\author[ ]{Sicong Leng\textsuperscript{1,*}, Yang Zhou\textsuperscript{2,*,†}, Mohammed Haroon Dupty\textsuperscript{3}, \\
Wee Sun Lee\textsuperscript{3}, Sam Conrad Joyce\textsuperscript{4}, Wei Lu\textsuperscript{1}}

\affil[1]{StatNLP Research Group, Singapore University of Technology and Design}
\affil[2]{Institute of High Performance Computing (IHPC), A*STAR Singapore}
\affil[3]{School of Computing, National University of Singapore}
\affil[4]{Meta Design Lab, Singapore University of Technology and Design}
% \affil[ ]{\texttt{sicong\_leng@mymail.sutd.edu.sg}}
 \affil[ ]{\texttt{\{sicong\_leng,sam\_joyce,luwei\}@sutd.edu.sg}}
 \affil[ ]{\texttt{zhou\_yang@ihpc.a-star.edu.sg}\ \ \texttt{\{dmharoon,leews\}@comp.nus.edu.sg}}
% \affil[ ]{\texttt{\{dmharoon,leews\}@comp.nus.edu.sg}}

\maketitle
\blfootnote{$^*$ Equal contribution † Most work done at NUS}
% \blfootnote{† Most work done at NUS.}

\begin{abstract}
%Engineering a design is a time-consuming, as well as an expertise-required process. It is exasperated by the less-informative user/client requirements in natural languages that the design needs to satisfy.
%As a representative, designing a floor plan, the structured interior building layout, is always protracted, involving multiple back-and-forths for expert designers to produce layouts that meet different user requirements. 
%To bridge the gap between user inputs and designs, 
We consider the task of generating designs directly from natural language descriptions, and consider floor plan generation as the initial research area. 
Language conditional generative models have recently been very successful in generating high-quality artistic images. However, designs must satisfy different constraints that are not present in generating artistic images, particularly spatial and relational constraints.
We make multiple contributions to initiate research on this task. First, we introduce a novel dataset, \textit{Tell2Design} (T2D), which contains more than $80k$ floor plan designs associated with natural language instructions. Second, we propose a Sequence-to-Sequence model that can serve as a strong baseline for future research. 
Third, we benchmark this task with several text-conditional image generation models. We conclude by conducting human evaluations on the generated samples and providing an analysis of human performance.
We hope our contributions will propel the research on language-guided design generation forward\footnote{\textcolor{black}{Code and dataset are available at \url{https://github.com/LengSicong/Tell2Design}.}}.

\end{abstract}

\section{Introduction}
Recently, text-conditional generative AI models \cite{nichol2021glide,saharia2022photorealistic,ramesh2022hierarchical,dhariwal2021diffusion,ho2022cascaded} have demonstrated impressive results in generating high-fidelity images. Such models generally focus on understanding high-level visual concepts from sentence-level descriptions, and the generated images are valued for looking realistic and being creative, thereby being more suitable for generating artwork. However, besides less constrained generation like artworks, generating designs that meet various requirements specified in natural languages is also much needed in practice \cite{stiny1980introduction,seneviratne2022dalle,zhang2022armani,wei2022hairclip}. 
In particular, a design process always involves interaction between users/clients, who define objectives, constraints, and requirements that should be met, and designers, who need to develop various solutions with domain-specific experiences and knowledge. 
For example, users may dictate their house design requirements in text and expect expert architects to perform the floor plan generation. 

\begin{figure}[t]
    \centering
    \includegraphics[width=0.95\columnwidth]{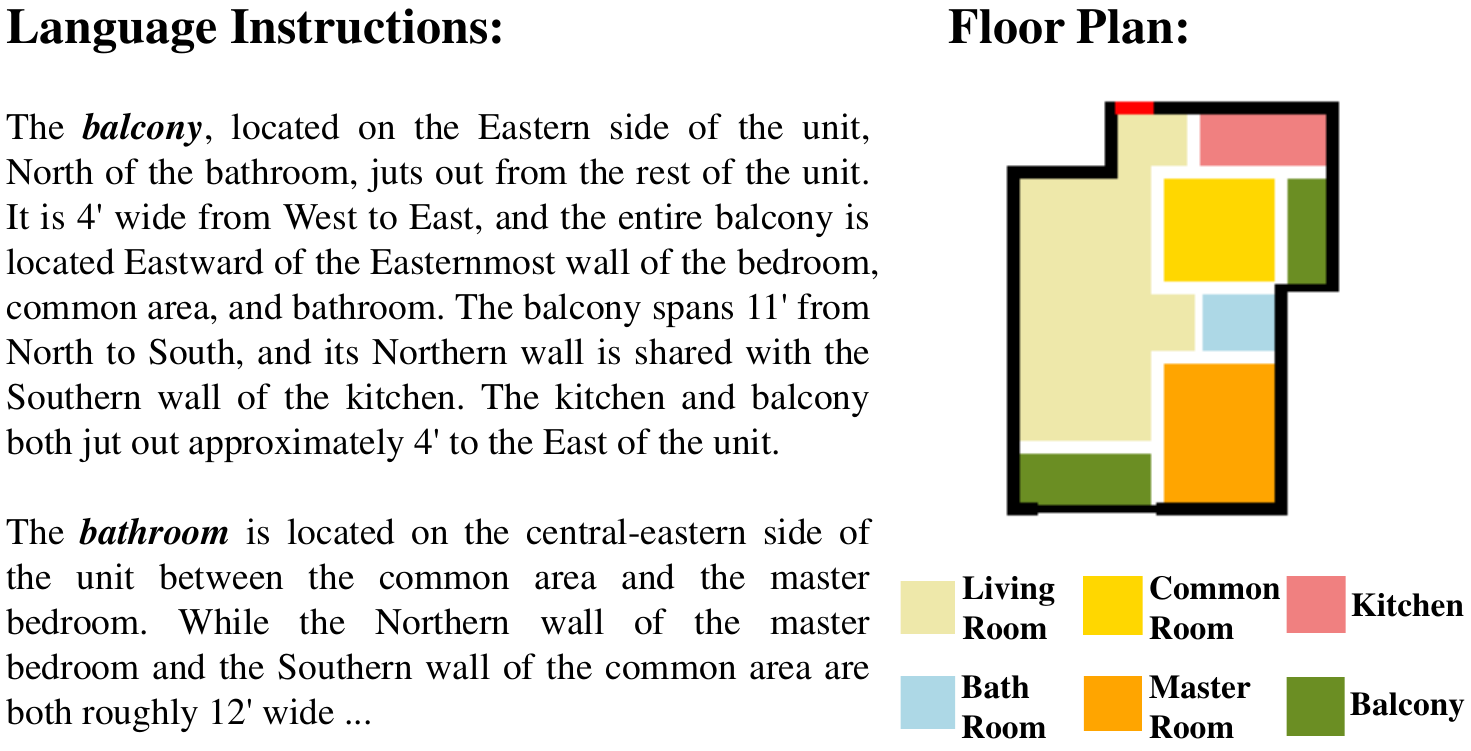}
    \caption{A data sample from our \textit{Tell2Design} dataset.} 
    \label{fig:intro}
\end{figure}

Previous research in layout generation aims to automate the process of layout design in different domains such as scientific documents, mobile UIs, indoor scenes, etc \cite{zhong2019publaynet,deka2017rico,song2015sun,janoch2013category,xiao2013sun3d,silberman2012indoor,cao2022geometry, wu2019data}. Most of them perform the generation either based on several hand-crafted constraints or by using unconstrained generation. In practice, it can be more convenient for users to indicate their preferences in natural language.
%that contain rich information and cannot be well-characterized by certain constraints. 

Among various design tasks, floor plan\footnote{Architectural floor plans, i.e., interior building layouts, are documents that indicate room types, room connections, room sizes, etc. They play a crucial role while designing, understanding, or remodeling indoor spaces \cite{liu2017raster}.} design, as shown in Figure~\ref{fig:intro} is of moderate complexity. However, it still intrinsically involves multiple rounds of communications between clients and designers for specifying requirements, and requires a high level of precision and alignment to detail.
AI systems that can learn to generate practically useful floor plan designs directly from natural languages will go a long way in reducing the protracted design process and making Generative AI directly usable for design by the end users.

To allow people without expertise to participate and further enhance the design process, we aim to enable users to design by ``telling'' instructions, with a specific focus on the floor plan domain as the initial area of research. This sets forth a new machine learning task where the model learns to generate floor plan designs directly from language instructions. However, this task brings up two technical challenges. First, a floor plan is a structured layout that needs three intrinsic components to be valid: (1) \textit{Semantics}, which describes the functionality of rooms (e.g., for living or bathing); (2) \textit{Geometry}, which indicates the shape and dimension of individual rooms; (3) \textit{Topology}, which defines the connectivity among different rooms \cite{pizarro2022automatic}. Second, these instructions are expressed in natural languages, which, besides the diversity of expressions, inherently suffer from ambiguity, misleading information, and missing descriptions for intrinsic components.

To address the above challenges, we make multiple contributions to initiate research on the task of \emph{language-guided floor plan generation}. First, we contribute a novel dataset, \emph{Tell2Design} (T2D), to the research community. The T2D dataset contains more than $80k$ real floor plans from residential buildings. Each floor plan is associated with a set of language instructions that describes the intrinsic components of every room in a plan. An example from the dataset is illustrated in Figure~\ref{fig:intro}. Second, we propose a Sequence-to-Sequence (Seq2Seq) approach as a solution to this task which also serves as a strong baseline for future research. Our approach is strengthened by a new strategy to explicitly incorporate the floor plan boundary constraint by transforming the outline into a box sequence. Third, in order to benchmark this novel task and evaluate our proposed approach, we implement strong baselines in text-conditional image generation on our T2D dataset and ask humans to perform the same task. The generation alignment with language instructions is evaluated both quantitatively and qualitatively. Finally, we discuss several future directions that are worth exploring based on our experimental results.

In summary, our main contributions are:
\squishlist
    \item We introduce a novel \emph{language-guided floor plan generation} task along with the T2D dataset consisting of both natural human-annotated and large-scale artificially generated language instructions (Section~\ref{section: dataset}). 
    \item We propose a new approach that formulates the floor plan generation task as a Seq2Seq problem (Section~\ref{section: method}). 
    \item We provide adequate quantitative evaluations on all baselines and qualitative analysis of human evaluations and performances (Section~\ref{section: exp}).
\squishend

\section{Related Work}
% In this paper, we focus on floor plan design, which is particularly important and widespread in real life. Existing floor plan generation techniques \cite{wu2018miqp,liu2013constraint,merrell2010computer,hua2016irregular,wu2019data,chen2020intelligent,chaillou2020archigan} are developed to produce layouts based on hand-crafted constraints, such as scene graphs and inequalities, that specify partial user preferences. However, users usually describe their desired layout in natural languages that is non-trivial to be well-characterized by certain constraints. We are not aware of any previous works that aim to generate floor plan designs directly from user language inputs.  To bridge this gap, we propose to directly generate floor plans from natural languages, providing a more flexible and user-friendly way to control the design process. 

% \textcolor{red}{Due to limited space, we provide detailed discussions on other similar tasks and their prevalent approaches in Appendix~\ref{sec: appendix related works}.}

\paragraph{Text-Conditioned Image Generation}
Image generation is a well-studied problem, and the most popular techniques have been applied for both unconditional image generation and text-conditional settings. Early works apply auto-regressive models \cite{mansimov2015generating}, or train GANs \cite{xu2018attngan,zhu2019dm,tao2020df,zhang2021cross,ye2021improving} with publicly available image captioning datasets to synthesize realistic images conditioned on sentence-level captions. Other works have adopted the VQ-VAE technique \cite{van2017neural} to text-conditioned image generation by concatenating sequences of text tokens with image tokens and feeding them into auto-regressive transformers \cite{ramesh2021zero,ding2021cogview,Aghajanyan2022CM3AC}. More recently, some works have applied diffusion models \cite{ho2020denoising,nichol2021improved,saharia2022image,dhariwal2021diffusion,ho2022cascaded,saharia2022palette,rombach2022high,nichol2021glide,saharia2022photorealistic,ramesh2022hierarchical} and received wide success in image generation, outperforming other approaches in fidelity and diversity, without training instability and mode collapse issues \cite{brock2018large,dhariwal2021diffusion,ho2022cascaded}. However, these models operate on extracting high-level visual concepts from the short text and produce artwork-like images that are expected to be realistic and creative, thereby not suitable for generating designs that must satisfy various user/client requirements.

\paragraph{Layout Generation}
Layout generation is essentially a design process that requires meeting domain-specific constraints, where the desirable layouts could be documents, natural scenes, mobile phone UIs, and indoor scenes. For example, PubLayNet \cite{zhong2019publaynet} is proposed to generate machine-annotated scientific documents with {\color{black}five different element categories \textit{text, title, figure, list, and table.}} RICO \cite{deka2017rico} is introduced to develop user interface designs for mobile applications, which contains \textit{button, toolbar, etc}. SUN RGB-D \cite{song2015sun} presents a combined scene-understanding task, including indoor scenes from three other datasets \cite{janoch2013category,xiao2013sun3d,silberman2012indoor}. Moreover, ICVT \cite{cao2022geometry} aims to produce advertisement poster layouts automatically, where the image background is given as input. 
The above methods are designed for different layout domains and cannot be directly applied to floor plan design. Moreover, none of them has considered generating the layout design directly from languages.

\paragraph{Floor Plan Generation}
Several methods have been proposed to generate floor plan designs automatically \cite{wu2018miqp,liu2013constraint,merrell2010computer,hua2016irregular,wu2019data,chen2020intelligent,chaillou2020archigan}. Most of these methods generate floor plans conditioned on certain constraints, such as room types, adjacencies, and boundaries. For example, \citet{merrell2010computer} generate buildings with interior floor plans for computer graphics applications using Bayesian networks without considering any human preferences. \citet{liu2013constraint} present an interactive tool to generate desired floor plan following a set of manually defined rules. \citet{hua2016irregular} particularly focus on generating floor plans with irregular regions. \citet{wu2018miqp} cast the generation as a mixed integer quadratic programming problem where some floor plan components are formulated into a set of inequality constraints. More recently, \citet{wu2019data} propose a CNN-based method to determine the location of different rooms given boundary images as a constraint. \citet{chen2020intelligent} provide a small amount of template-based artificial verbal commands and manually parses them to scene graphs for guiding the generation. 

In summary, existing methods represent the intrinsic components of floor plans in several specific formats as generation constraints. Some formats are straightforward, such as boundary images, but they only specify limited constraints and lead to less controllable generation. Other formats, such as scene graphs and inequalities, can incorporate more information but require specific domain knowledge and extra-human efforts in pre-processing. 
We instead provide a unified and natural way of conditioning the floor plan generation with a set of language instructions that is much more flexible and user-friendly to characterize floor plans with various constraints. 

\section{Tell2Design Dataset}
\label{section: dataset}

In this section, we introduce how we construct our T2D dataset, followed by the data analysis and a discussion of the main dataset challenges.

\subsection{Task Definition}
Given a set of language instructions describing a floor plan's intrinsic components, our aim is to generate reasonable 2D floor plan designs that comply with the provided instructions. 
%An example is illustrated in Figure~\ref{fig: RPLAN3}.

% \begin{figure}[]
%     \centering
%     \includegraphics[scale=0.2]{imgs/t2d example3.png}
%     \caption{A floor plan example with artificial and human language instructions for the master room and balcony, respectively. Keywords for \textit{Semantics} are highlighted in green, \textit{Geometry} in red, and \textit{Topology} in yellow.} 
%     \label{fig: RPLAN3}
% \end{figure}

\paragraph{Input \& Output}
For each data sample, the input is a set of natural language instructions that characterize the key components of the corresponding floor plan design, which include: 
(1) \textit{Semantics} specifies the type and functionality of each room. For example, a room as \textit{Kitchen} is for cooking.
(2) \textit{Geometry} specifies the shape and dimension of each room. For residential buildings, it involves the room's general orientation (e.g., the north, south, northeast, southwest), area in square feet, aspect ratio, etc. 
(3) \textit{Topology} describes the relationships among different rooms. It can be divided into three categories: relative location, connectivity, and inclusion\footnote{As a result, language instructions specifying the above features for a floor plan lead to a document-level description in natural language. We compare the T2D dataset with several document-level NLP datasets in Appendix~\ref{sec: appendix dataset analysis}.}.
The desirable output is a structured interior layout that aligns with the input language instructions.

\subsection{Floor Plan Collection}
We use floor plans from RPLAN\footnote{\url{http://staff.ustc.edu.cn/~fuxm/projects/DeepLayout/index.html}} \cite{wu2019data} to construct our Tell2Design dataset. We remove floor plans with rarely-appeared rooms and merge similar room types such as \textit{Second Room} and \textit{Guest Room}. As a result, $8$ different room types (i.e., common room, bathroom, balcony, living room, master room, kitchen, storage, and dining room) and $80,788$ floor plans are selected for collecting language instructions. Each floor plan is converted into a $256 \times 256$ image where different pixel values indicate different room types, from which we extract room-type labels and bounding boxes of each room to construct our dataset. 

\begin{table}[t]
\begin{center}
    \centering
    \small
\resizebox{0.95\columnwidth}{!}{%
    \begin{tabular}{lrr}
    \toprule
                                & Human         & Artificial   \\ \midrule
    Avg. \# words per instance  & 200.30        & 260.47             \\
    Avg. \# sent. per instance  & 11.89         & 23.46              \\ 
    Avg. \# words per room      & 29.48         & 38.44              \\ 
    Avg. \# sent. per room      & 1.75          & 3.46               \\
    \bottomrule
    \end{tabular}
    }
\end{center}
\caption{Language instruction statistics.} 
\label{tab: language stats}
\end{table}
\subsection{Language Instruction Collection}
\label{data annotation}
\paragraph{Human Instructions}
To collect real human language instructions, we hire crowdworkers from Amazon Mechanical Turk (MTurk)\footnote{\url{https://www.mturk.com/}} and ask them to write a set of instructions for each room according to a given floor plan image. The requested instructions should reflect the \textit{Semantic, Geometric,} and \textit{Topological} information of the floor plan, such that designers could ideally reproduce the floor plan layout according to the instructions. In particular, \textcolor{black}{turkers are encouraged to include (but are not limited to) attributes such as room types, locations, sides, and relationships in their instructions. The definitions of these attributes are given as follows: 
The room type (e.g., \texttt{bathroom and kitchen}) specifies the functionality of a room. 
The room location specifies the global location of a room in the floor plan and can be described by phrases such as ``\texttt{north side}'' and ``\texttt{southeastern corner}''. 
The room sides specify the length and width of a room (e.g., ``\texttt{8 feet wide and 10 feet long}''). 
The room relationships specify the relative position of a room with other rooms such as ``\texttt{next to}'', ``\texttt{between}'', and ``\texttt{opposite}''
\footnote{\textcolor{black}{To mimic the real-world scenarios, we do not provide any bounding box information and ask crowdworkers to make rough estimations of the room size from the given floor plan image only. We also do not restrict the format of text descriptions or require all the above attributes to be mentioned, leading to unstructured and diverse instructions.}}.}

\textcolor{black}{Due to the noisy nature of crowdsourcing annotations, we discard some low-quality annotations to ensure the overall quality of our datasets. To this end, we manually review each annotation and discard human instructions with: 
(1) incoherence, grammatical errors;
(2) insufficient attributes;
or (3) irrelevance to the given floor plan.}
As a result, we collect human instructions for $8,220$ floor plans, and $5,051$ of them are finally accepted after manual assessment to construct our dataset\footnote{Our human instruction collection involves $5,109$ different workers with $1,723$ working hours, \textcolor{black}{and each worker receives full compensation in alignment with the established standards of MTurk.}}. 

%is fully compensated according to MTruk standards.
%and costs USD $3,954$ with $1,723$ working hours. 
%More details on the human instruction collection can be found in Appendix~\ref{sec:appendix}. 

\paragraph{Artificial Instructions}
In addition to the human-written instructions, we also generate language instructions artificially for the remaining $75,737$ floor plans from pre-defined templates. To ensure that the artificial instructions are as informative as human-written ones and include all the required components, we ask 5 educated volunteers with natural language processing (NLP) backgrounds to write language instructions for each room that appeared in the given floor plan. We then summarize their instructions into multiple templates and ask expert architectural designers for proofreading.
%As a result, summarized templates contain descriptive text for individual rooms (e.g., room type, location, size, aspect ratio, and relationships). 
Hence, each instruction template is ensured to be informative, grammatically correct, and coherent. 
%We show the example of an artificial template in Appendix~\ref{sec:appendix}. 

In summary, our T2D dataset consists of $5,051$ human-annotated and $75,737$ artificially-generated language instructions\footnote{More details on the language instruction collection can be found in Appendix~\ref{sec: appendix dataset collection}.}.
\subsection{Data Analysis}
\label{section: data analysis}
In this section, we analyze various aspects of Tell2Design to provide a more comprehensive understanding of the dataset.

\paragraph{Language Instructions}

Table~\ref{tab: language stats} shows the statistics of the language instructions in our dataset. 
For each floor plan, the human instructions are organized in nearly 11 sentences consisting of $200$  words on average.
%For each floor plan, the average number of words of the language instructions is around $200$ to $300$ contained in nearly 11 sentences. 
This includes around 30 words used to describe each room in more than $2$ sentences. The artificially generated instructions follow a similar pattern with slightly more words.

\begin{figure}[t]
    \centering
    \includegraphics[scale=0.32]{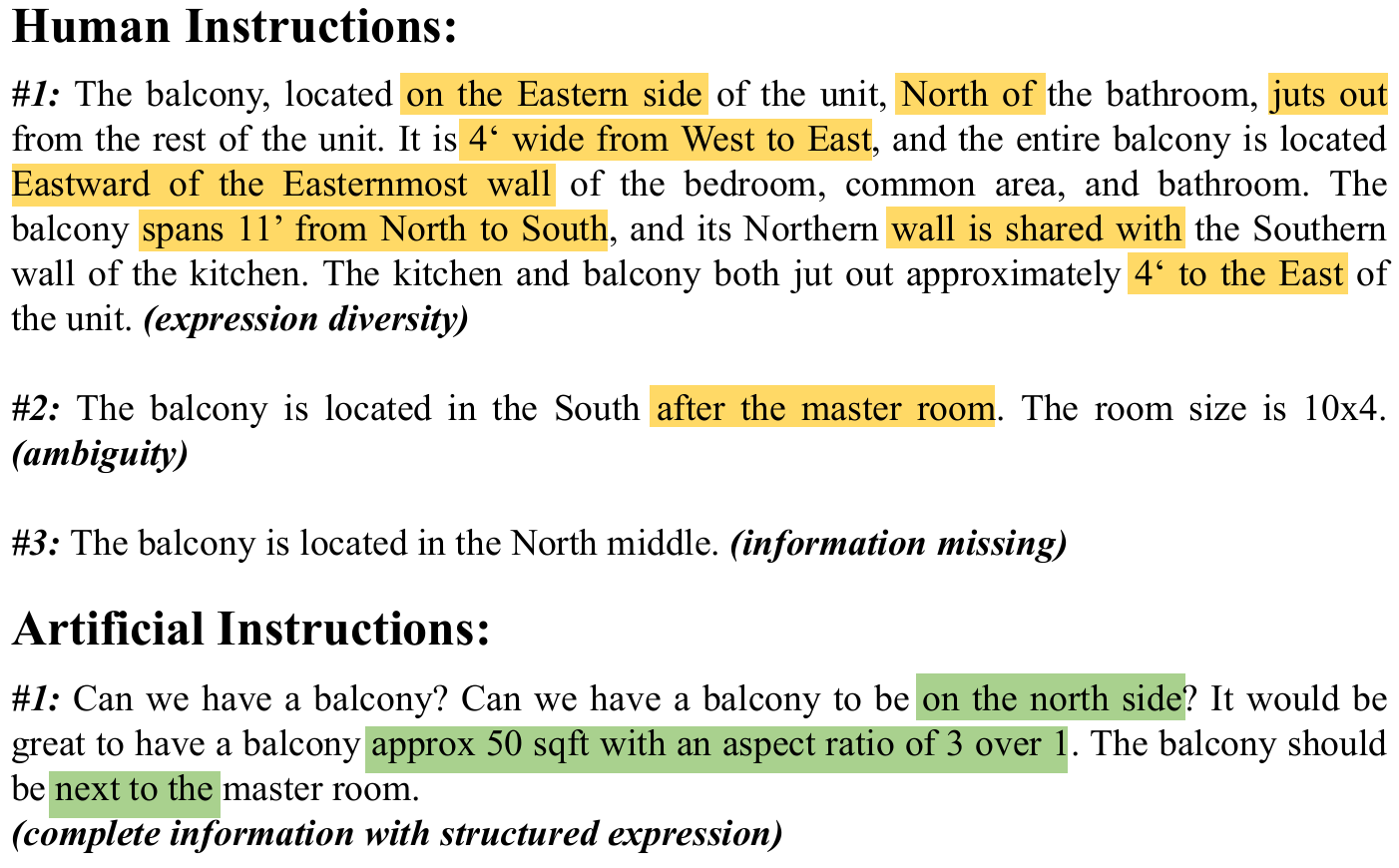}
    \caption{Human instructions vs. artificial instructions.} 
    \label{fig: language analysis}
\end{figure}

To show the connections and differences between human and artificial instructions, we compare them for the same room type, \textit{Balcony}, in Figure~\ref{fig: language analysis}. Artificial instructions always exhibit complete information, including all three key components of a floor plan. They are also formatted in a structured expression, such as ``\texttt{on the ** side}'', ``\texttt{** sqft with an aspect ratio of **}'', and ``\texttt{next to}''. However, human instructions are more diverse in expression but suffer from ambiguity and missing components.

\paragraph{Dataset Comparison} 
In order to see our T2D dataset in perspective, we note its main differences with respect to other related datasets used for similar generation tasks\footnote{We provide detailed comparisons with tables in Appendix~\ref{sec: appendix dataset analysis}.}. T2D differs from other datasets in several perspectives: (1) T2D is the first large-scale dataset that aims to generate designs (i.e., floor plans) from direct user input natural language; (2) T2D has much longer text annotations (i.e., $256$ words per instance) compared with other text-conditional generation datasets; (3) All text in T2D is written by humans or generated artificially, instead of being crawled from the internet. 

\subsection{Dataset Challenges}
In this section, we discuss three main challenges of our collected T2D dataset. We hope this dataset can facilitate the research on both design generation and language understanding. 

\textbf{Design Generation under Constraints} The first challenge is to perform the design generation under much stricter constraints compared with artwork-like text-conditional image generation. Most works in text-conditional image generation operate on generating realistic and creative images that align with the main visual concepts represented by the short input text. 
%Most works in text-conditional image generation operate on a text encoder that can encode the input short descriptions to condensed latent representations and an image decoder for generating realistic and creative images that align with the main visual concepts represented by the input text. 
However, creating a design from languages has much stricter requirements on precision and alignment to text details. In particular, the generated floor plan design should comply with constraints such as room type, location, size, and relationships, which are specified by users using natural languages.
Our main results in Section~\ref{section: exp} comparing different baselines demonstrate that existing text-conditional image generation techniques fail to follow detailed user requirements on this design task.
%We directly evaluate those detailed partial alignments in Section~\ref{subsec: analysis}.

\textbf{Fuzzy \& Entangled Information} The second challenge is to understand the big picture of the entire floor plan from document-level unstructured text with fuzzy and entangled information. Besides the general abilities required for language understanding, such as entity recognition, coreference resolution, relation extraction, etc, models also need to collaborate with fuzzy individual room attributes and reason over entangled relationships among different rooms to understand the entire floor plan. Specifically, one language instruction usually either specifies fuzzy descriptions for a room's \textit{Semantic} and \textit{Geometric} information such as ``\texttt{on the north side}'' and ``\texttt{at the southeast corner}'', or indicates the relationship of one specific room with others like ``\texttt{next to}'' and ``\texttt{between}''. The provided information in such instructions is coarse and relative, rather than complete and precise information like numerical coordinates. As a result, to determine the location of all rooms and design a reasonable floor plan, models must collaborate with fuzzy and entangled information residing in multiple instructions, and incorporate the boundary information. Human evaluations in Section~\ref{subsec: analysis} demonstrate that room relationships described in language instructions are the most challenging component to be understood and aligned with.

\textbf{Noisy Human Instructions} The third challenge comes from the ambiguous, incomplete, or misleading information in human instructions. As introduced in Section~\ref{data annotation}, the artificial instructions are template-based so that they always contain precise and coherent information. However, for human-written language instructions, ambiguous or noisy information always exists. For example, during human instruction collection, workers are asked to write natural sentences estimating some numeric-related attributes like room size and aspect ratio referring to the floor plan image, which may sometimes be inaccurate.
Moreover, as previously discussed in Figure~\ref{fig: language analysis}, other than the expression diversity, human instructions also exhibit ambiguous phrasing and incomplete information. It is thus more challenging for models to retrieve accurate, complete, and consistent information from human instructions.

\section{T2D Model}
\label{section: method}
In this section, we propose a simple yet effective method for \textit{language-guided floor plan generation}. Unlike \textcolor{black}{existing floor plan generation methods \cite{wu2019data,chen2020intelligent}} that use a regression head to generate the bounding box of each room one at a time, we cast the floor plan generation task as a Seq2Seq problem under the encoder-decoder framework, where room bounding boxes are re-constructed into a \textit{structured} target sequence. This way, our method can easily deal with various lengths of instructions for floor plans with different numbers of rooms.
%and generate all the desirable bounding boxes simultaneously. 
%In addition, we further propose a novel approach to incorporate the floor plan outline/boundary information by transforming floor plan outlines with irregular shapes into a set of bounding boxes. Such outline boxes are then represented into sequences of tokens and combined with language instructions for training, leading to a significant performance boost.

\subsection{Target Sequence Construction}
\label{subsec: target seq}
Recall that our aim is to generate a floor plan layout from language instructions, where each room can be represented by a room-type label (e.g., bathroom and kitchen) and a bounding box. One bounding box can be determined by four values $(x, y, h, w)$, which indicate the $x$ and $y$ coordinate of the center point, height ($h$), and width ($w$), respectively. To solve language-guided floor plan generation as a Seq2Seq problem, we treat the instructions as the input sequence and consider bounding boxes of rooms as the target sequence. Specifically, each of the continuous values $(x, y, h, w)$ is discretized into integers between $[0,255]$, and the room type is given by the plain text in natural language, so that they can be naturally represented as a sequence of tokens. The target sequence is then constructed by grouping the room type and the bounding box together with certain special tokens. For example, the target sequence for a \textit{Balcony} with the bounding box $(87, 66, 18, 23)$ is given as follows:
\begin{align*}
    & \texttt{[ Balcony | x coordinate = 87 | y coordi} \\
    &\texttt{nate = 66 | height = 18 | width = 23 ]},
\end{align*}
where the special tokens ``\texttt{[}'' and ``\texttt{]}'' are used to indicate the start and end of the target sequence for one room and ``\texttt{|}'' is used to separate different target components. We have also added semantic prefixes such as ``\texttt{x coordinate =}'' and ``\texttt{height =}'' before the values of bounding boxes to assist the target sequence generation. Finally, we concatenate the target sequences of all the rooms in a floor plan and add an \texttt{<eos>} token at the end to indicate the end of the overall target sequence.

\subsection{Boundary Information Incorporation}
\begin{figure}
    \centering
    \includegraphics[scale=0.4]{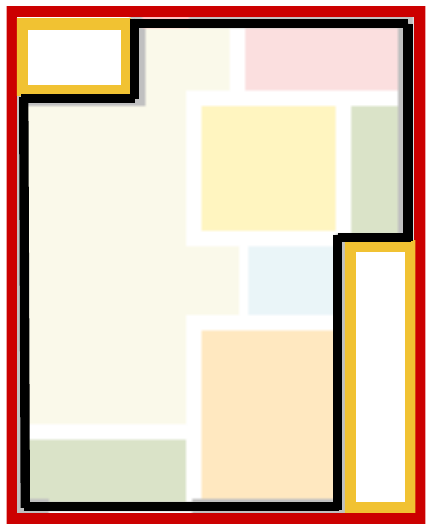}
    \caption{An illustration of how to transform visual floor plan boundaries into boxes.} 
    \label{fig: boundary process}
\end{figure}
The outline/boundary of a floor plan is one of the most important constraints in floor plan generation, which directly affects where each room should be placed and how different rooms should be aligned with the floor plan boundary. However, it is non-trivial to incorporate such boundary information into floor plan generation. Previous methods either fail to take the floor plan outline into account \cite{wu2018miqp,liu2013constraint,merrell2010computer,hua2016irregular,chen2020intelligent} or only consider the boundary image, ignoring all other constraints \cite{wu2019data,chaillou2020archigan}, leading to less controllable floor plan design.

In this work, we propose a novel approach to incorporate boundary information by representing the irregular outline as a set of boxes. The idea is to encode the boundary information by an enclosing box that is the minimum bounding region containing the entire floor plan and several exterior boxes that are inside the enclosing box but excluded from the floor plan. Figure~\ref{fig: boundary process} illustrate how the floor plan boundary can be characterized by the enclosing (in red) and exterior boxes (in yellow). This way, we have an enclosing box represented by $(x^{en},\:y^{en},\:h^{en},\:w^{en})$ and $M$ exterior boxes by $(x_i^{ex},\:y_i^{ex},\:h_i^{ex},\:w_i^{ex})$. Then we adopt a similar strategy in Section~\ref{subsec: target seq} to represent the enclosing and exterior boxes in a sequence as follows:
% \begin{align*}
% & [+,\:x^{en},\:y^{en},\:h^{en},\:w^{en},\:-,\:x_1^{ex},\:y_1^{ex}, \\ 
% & h_1^{ex},\:w_1^{ex},...,\:-,x_M^{ex},\:y_M^{ex}, h_M^{ex},\:w_M^{ex}] ,
% \end{align*}
\begin{align*}
& \texttt{+ $x^{en}$ $y^{en}$ $h^{en}$ $w^{en}$ - $x_1^{ex}$ $y_1^{ex}$} \\ 
& \texttt{$h_1^{ex}$ $w_1^{ex}$ ... - $x_M^{ex}$ $y_M^{ex}$ $h_M^{ex}$ $w_M^{ex}$} ,
\end{align*}
where the coordinates of the enclosing and exterior box are following the tokens ``\texttt{+}'' and ``\texttt{-}'', respectively.

Finally, the above sequence is added after the input language instructions for training. Our experimental results in Section~\ref{section: exp} show that the proposed boundary information incorporation strategy is effective in enhancing our Seq2Seq method to generate valid rooms that align well with the floor plan boundary.
%generating valid rooms that align well with the boundary and significantly improve the performance of floor plan generation.

\subsection{Architecture, Objective and Inference}
Treating the target sequences that we construct from floor plans as a text sequence, we turn to recent architectures and objective functions that have been effective in Seq2Seq language modeling.

\paragraph{Architecture} We use the popular Transformer-based \cite{vaswani2017attention} encoder-decoder structure to build our Seq2Seq model for floor plan generation. The model is initialized by a pre-trained language model T5 \cite{raffel2020exploring} for better language understanding abilities\footnote{We have also tried other pre-trained language models like Bart \cite{lewis2020bart}, and preliminary experiments indicate that initializing our model with T5 leads to better performances.}.

\paragraph{Objective} Similar to language modeling, our T2D model is trained to predict the next token, given an input sequence and preceding tokens, with a maximum likelihood objective function, \ie,
\begin{equation}
    \max_\theta \sum_{j=1}^L \log P_\theta \left(\tilde{\bm{y}}_j \mid \bm{x}, \bm{y}_{1: j-1}\right) {,}
\end{equation}
where $\bm{x}$ is a set of instructions in natural language concatenated with the previously defined boundary sequence, $\bm{y}$ is the target bounding box sequence, and $L$ is the target sequence length.

\paragraph{Inference} At inference time, we sample\footnote{There are several common sampling strategies like \textit{Greedy Search, Beam Search, and Nucleus Sampling} \cite{holtzman2019curious}. We apply \textit{Greedy Search} since it leads to better generation quality in our preliminary experiments.} tokens one by one from the model likelihood, \ie, $P\left(\tilde{\bm{y}}_j \mid \bm{x}, \bm{y}_{1: j-1}\right)$. The sequence generation ends once the \texttt{<eos>} token is sampled, and it is straightforward to parse the target sequence into predicted floor plans.

\section{Experiments}
\label{section: exp}
\subsection{Baselines}
Since our T2D dataset is the first to consider language-guided floor plan generation, existing layout generation methods are not applicable to this task. 
To further illustrate the challenge of the design generation task and the difference with the existing text-conditional image generation problem, we adapt several state-of-the-art text-conditional image generation methods as baselines for comparison. In particular, we compare our method with the following:
%Therefore, we represent floor plans as images and adapt several state-of-the-art text-conditional image generation methods as baselines for comparison. In particular, we compare our method with the following:
%Obj-GAN\cite{li2019object}, CogView\cite{ding2021cogview}, and Imagen\cite{saharia2022photorealistic}, which are trained with text-image pairs.
\squishlist
\item Obj-GAN \cite{li2019object} is an object-driven attentive generative adversarial network that follows a two-step generation process. %where the semantic layout is generated from text descriptions before synthesizing the final image. 
We apply the first-step box generator module, which takes the language as input and generates target objects' bounding boxes (with class labels).

\item CogView \cite{ding2021cogview} applies pre-trained VQ-VAE to transform the target image into a sequence of image tokens. Then the text and image tokens are concatenated together and fed to a Transformer decoder (\ie, GPT \cite{brown2020language,radford2019language}) to generate text-conditional images.

\item Imagen \cite{saharia2022photorealistic} is one of the state-of-the-art text-to-image generation models that build upon both large language models (e.g., T5) for text understanding and diffusion models for high-fidelity image generation.
\squishend

\subsection{Experimental Settings}
\paragraph{Model Training}
For model training, we consider a Warm-up + Fine-tuning pipeline \cite{goyal2017accurate}, where the model is first warmed up on $75,737$ artificial instructions, and then fine-tuned on $2,743$ human instructions.
%\footnote{Warm-up on $75,737$ floor plans with artificial instructions, fine-tune on $2,743$ human instructions, and test on another $2,308$ human instructions.}.
%Our T2D dataset consists of $5,051$ human-annotated and $75,737k$ artificially-generated instructions. 
To evaluate how floor plan generation methods generalize to unseen instructions, we use the remaining $2,308$ human instructions as the test set, such that there is no overlapping between annotators of the training set and the test set\footnote{We provide more implementation details in Appendix~\ref{sec: appendix implementation details}.}.
%randomly split the human-written instructions into a training set and a test set with similar sizes such that the annotators of training instructions will not appear in the test set.  

\paragraph{Evaluation Metrics}
For testing, we use macro and micro Intersection over Union (IoU) scores between the ground-truth (GT) and generated floor plans at pixel level as the evaluation metrics, whose definitions are given as follows:
\begin{equation*}
    \text{Micro IoU} = \frac{\sum_{r=1}^{R} I_r}{\sum_{r=1}^{R} U_r}, \text{Macro IoU} = \frac{1}{R}\sum_{r=1}^{R}\frac{I_r}{U_r}, 
\end{equation*}
% \begin{align*}
%     \text{Micro IoU} &= \frac{\sum_{r=1}^{R} I_r}{\sum_{r=1}^{R} U_r} \\
%     \text{Macro IoU} &= \frac{1}{R}\sum_{r=1}^{R}\frac{I_r}{U_r}, 
% \end{align*}
where the $I_r$ and $U_r$, respectively, denote the intersection and union of the ground-truth and predicted rooms labeled as the $r$-th room type in a floor plan. $R$ is the total number of room types.
Macro IoU calculates the average IoU over different types of rooms, and Micro IoU calculates the global IoU by aggregating all rooms.

% Specifically, for box-based approaches such as Obj-GAN and our Seq2Seq method, the IoU scores are directly computed using predicted and ground-truth bounding boxes. For image-based approaches (e.g., CogView and Imagen), we first adjust all pixel values in generated images to their nearest value that has a semantic meaning in our T2D dataset (e.g., $(0,255,255)$ represents the color for the bathroom), and then compute the maximized IoU scores in pixel level by shifting the floor plan central point in the generated image.

Since Obj-GAN and our T2D model generate bounding boxes rather than images, we use a simple strategy to transform the outputs of Obj-GAN and the T2D model into images without any further refinement for a fair comparison. Specifically, we paint each room in descending order in terms of the total area of the room type\footnote{Total area of the room type is computed by adding up the specific room type area across all floor plans in our dataset. This gives us the following order: living room, common room, master room, balcony, bathroom, kitchen, storage, dining room.} and different colors refer to different room types. Previous colors will be replaced by the subsequent paintings if there is an overlapping between bounding boxes (rooms). For image-based approaches (e.g., CogView and Imagen), we compute the maximized IoU scores by shifting the floor plan central point in the generated image.

\subsection{Main Results}
\label{subsec: results}

\begin{table}[t]
\begin{center}
\centering
\small
\centering
\resizebox{0.88\columnwidth}{!}{%
    \begin{tabular}{lrr}
    \toprule
    \textbf{Models}  & \multicolumn{1}{r}{\space Micro IoU}            & \multicolumn{1}{r}{  Macro IoU}                        \\ \midrule
    \multicolumn{3}{c}{Training on artificial instructions only}                                                  \\ \midrule
    \multicolumn{1}{l}{Obj-GAN} & 15.74                & 11.12                                   \\
    \multicolumn{1}{l}{CogView} & 10.01                & 8.31                                 \\
    \multicolumn{1}{l}{Imagen}  & \textbf{14.74 }               & \textbf{15.57}                              \\ 
    \multicolumn{1}{l}{T2D (w/o bd)}  & 6.46                & 4.01                                \\ 
    \multicolumn{1}{l}{T2D}      & 9.13                & 6.06                                   \\\midrule
    \multicolumn{3}{c}{Training on human instructions only}                                                       \\ \midrule
    \multicolumn{1}{l}{Obj-GAN} & 10.72                & 8.29                                    \\
    
    \multicolumn{1}{l}{CogView} & 13.48                & 11.26                                 \\
    \multicolumn{1}{l}{Imagen}  & 9.29                 & 6.64                                   \\
    \multicolumn{1}{l}{T2D (w/o bd)}  & 32.22                & 26.24                                \\
    \multicolumn{1}{l}{T2D}      & \textbf{42.93}       & \textbf{38.48}                             \\\midrule
    \multicolumn{3}{c}{Warm up on artificial + fine-tune on human}                                     \\ \midrule
    \multicolumn{1}{l}{Obj-GAN} & 10.68                & 8.44                                  \\
    
    \multicolumn{1}{l}{CogView} & 13.30                & 11.43                               \\
    \multicolumn{1}{l}{Imagen}  & 12.17                & 14.96                            \\ 
    \multicolumn{1}{l}{T2D (w/o bd)}  & 35.95                & 29.95                                \\
    \multicolumn{1}{l}{T2D}      & \textbf{54.34}       & \textbf{53.30}                       \\\bottomrule
    \end{tabular}
    }
\end{center}
\caption{IoU scores between ground-truth and generated floor plans for the T2D model and other baselines.} 
\label{tab: results}
\end{table}

Table~\ref{tab: results} shows the floor plan generation results on the T2D dataset, where T2D (w/o bd) indicates the T2D model without incorporating boundary information\footnote{We provide baseline generation samples in Appendix~\ref{sec: appendix baseline}.}. The T2D model achieves the highest IoU scores with a micro IoU of $54.34$ and a macro IoU of $53.30$, outperforming other baselines by a large margin. These can be attributed to our Seq2Seq model in controlling the target box sequence generation based on salient information extracted from language instructions.
%extracting floor plan information from instructions and controlling the target sequence generation for the final layout. 
In contrast, text-conditional image generation methods fail to perform well. 
%This is probably because generating layouts as images rather than bounding boxes has a much larger output space and is more difficult to be controlled with detailed language instructions.
This is probably because those models are designed to generate artwork-like images with high-level visual concepts from the short text, instead of following multiple instructions with various constraints for a specific design.

When training only on artificial instructions while testing on human-written ones, our method cannot perform well. This indicates there is a language distribution gap between artificial and human instructions. 
Nevertheless, when artificial instructions are used for warming up before training on human instructions, the performance of our method is significantly improved with over $10$ IoU scores increment.
This suggests that despite the language gap, artificial and human instructions are mutually beneficial data portions during training.

In addition, in all the training settings, representing the floor plan boundary as a sequence of boxes consistently improves the performance of our Seq2Seq approach. This demonstrates that this strategy could be one of the possible solutions to incorporate the floor plan boundary.
%This demonstrates that representing the boundary as a set of boxes can effectively help our Seq2Seq method incorporate the boundary information and produce better floor designs.

\subsection{Result Analysis}
\label{subsec: analysis}
It is worth noting that the quantitative results \textit{indirectly} evaluate how well the generated floor plans align with the language instructions since IoU scores essentially measure the overlap between generated and ground-truth floor plan layouts. Due to the complexity of our task, it is possible for the same language instruction to map to multiple floor plan designs. Therefore, a low IoU score does not necessarily mean a bad generation.

\begin{table}[t]
\begin{center}
\centering
\small
\resizebox{0.9\columnwidth}{!}{%
\begin{tabular}{lrr}
\toprule
\textbf{Alignment}           & GT ratings  & T2D ratings \\ \midrule
Room type           & 4.99 & 4.71 \\
Room location       & 4.86 & 3.67 \\
Room size           & 4.75 & 3.89 \\
Relationships & 4.89 & 3.65 \\ \midrule
Meet all \%         & 85\% & 38\% \\ \bottomrule
\end{tabular}
}
\end{center}
\caption{Human evaluation results.} 
\label{tab: human results}
\end{table}

\paragraph{Human Evaluations}
% \paragraph{Human evaluations.}
To \textit{directly} evaluate the alignment between generated floor plans and language instructions, we conduct human evaluations on a subset of the T2D test set, which consists of $100$ randomly sampled instructions written by different annotators. For this purpose, we invite 5 volunteers with NLP backgrounds to evaluate the degree of alignment between source language instructions and target floor plans. 

Specifically, we consider four partial alignment criteria in terms of room types, locations, sizes, and relationships. Each volunteer is asked to provide four ratings on a scale of $1$ to $5$, according to the above-mentioned alignment criteria, respectively\footnote{For example, a rating of $r$ with respect to room locations indicates that $(r-1) \times 20\%$ to $r \times 20\%$ rooms in the floor plan follow their location specifications in the instructions.}. Besides, we also consider global alignment and ask our volunteers to justify whether the generated floor plan meets \textit{all} the specifications in the instructions. We perform the above subjective evaluations for both T2D-generated and ground-truth floor plan designs.
%In addition to the floor plan generated by our T2D method, we also perform the above subjective evaluations for the ground-truth floor plan designs.

Table \ref{tab: human results} shows the human evaluation results. As can be seen, ground-truth floor plans get high ratings for all the partial alignment criteria, and $85\%$ of them meet all the requirements specified in the instructions. This indicates our dataset contains high-quality human instructions that align well with the ground-truth floor plan designs. On the other hand, our T2D model receives no rating less than $3.5$, indicating that at least $50\%$ rooms with respect to their locations, sizes, and relationships can be correctly predicted. 
%Besides, up to $38\%$ of generated floor plans satisfy all the language specifications, which again demonstrates the effectiveness of our method in generating plausible floor plans. 
However, our method still has a gap with ground truth designs, especially in room location and relationships, which indicates the potential for improvements.
%\footnote{Please note that it is expected that even the ground-truth floor plans do not perfectly align with the instructions, which can be attributed to typos, ambiguous descriptions, or missing information in human-written instructions.}

% \begin{figure}[]
%     \centering
%     \includegraphics[scale=0.4]{imgs/human performance.png}
%     % \caption{Comparisons between T2D and human performance.} 
%     \caption{T2D vs. human performance.} 
%     \label{fig: human performances}
% \end{figure}
\begin{table}[]
\begin{center}
\centering
\small
\resizebox{0.75\columnwidth}{!}{%
\begin{tabular}{lrr}
\toprule
\textbf{Models}            & Micro IoU & Macro IoU \\ 
\midrule
T2D               & 55.10     & 55.16     \\
Human             & 64.67     & 62.32    \\ 
\bottomrule
\end{tabular}
}
\end{center}
\caption{The T2D model vs. human performance.} 
\label{tab: human performance}
\end{table}

\begin{figure}[]
    \centering
    \includegraphics[scale=0.4]{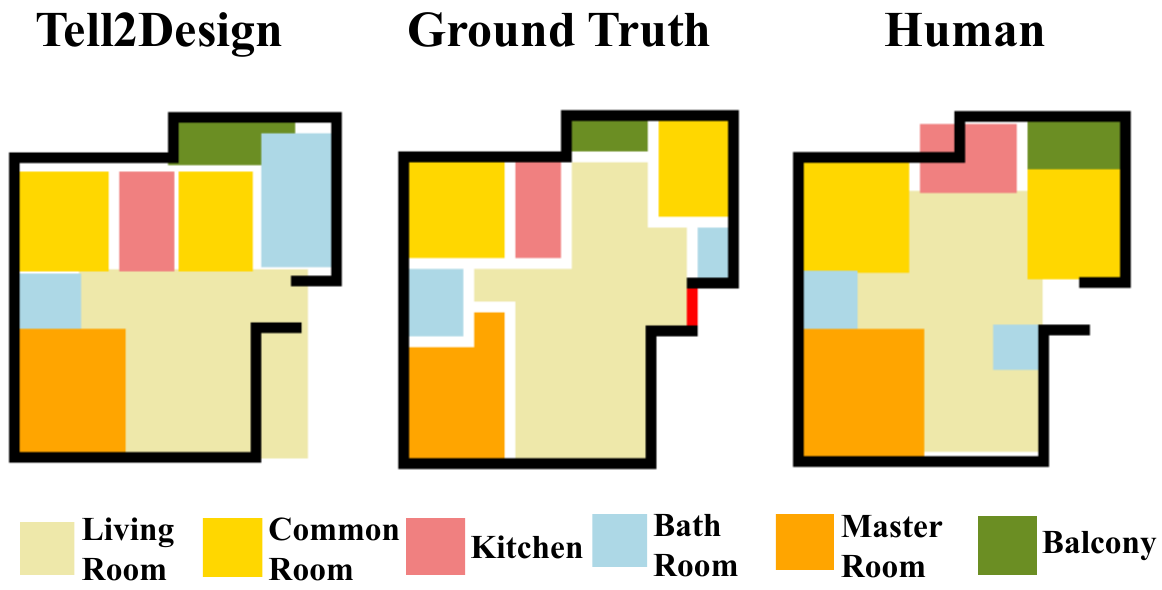}
    \caption{Floor plan layouts from our T2D model, the ground-truth, and human annotators based on the same language instructions. Although all of them satisfy the given instructions, they are different in detail, especially for the upper right corner of the floor plan.} 
    \label{fig: case study}
\end{figure}

\paragraph{Human Performance}
To study human performance on the T2D task, we further ask our volunteers to design floor plans on $100$ instances of the same subset used for human evaluations\footnote{Specifically, volunteers are asked to draw a bounding box for each room, given the floor plan outline and the language instructions.}.  Table~\ref{tab: human performance} reports the IoU scores for our T2D model and human performance. Humans generally achieve better IoU scores. However, even if human-generated floor plans intrinsically have much better alignment with the instructions, they only obtain around $63\%$ IoU scores with the ground truths. This exposes the nature of design diversity, \ie, a set of language instructions can map to multiple plausible floor plan designs. Figure~\ref{fig: case study} provides a real example that both our method and humans follow the same instructions\footnote{The language instructions for this sample are shown in Figure~\ref{fig: baseline} in Appendix~\ref{sec: appendix baseline}. } but generate different floor plans.

\section{Future Research}
%language-guided floor plan generation is a challenging task in many aspects. 
%Baselines, experimental results, and human analysis are presented as foundations for future research, and demonstrate the value of our proposed task and Tell2Design dataset. 
In the future, the following directions may be worth exploring to promote the performance or extend our task: 
%(1) How to effectively reason over and understand the entangled relationships among multiple rooms to generate floor plans with better alignment? 
(1) How to build robust language understanding models that can adapt to the presence of noise in human instructions or even locate and refine potentially inconsistent information? (2) How to explicitly incorporate the nature of design diversity and develop techniques for diverse floor plan design? (3) How to extend the language-guided floor plan generation task to more domains or more practical but challenging scenarios, where designs should be refined according to feedback from users/clients?
% \squishlist
% \item How to effectively reason over and understand the entangled relationships among multiple rooms to generate floor plans with better alignment?
% \item How to build robust language understanding models that can adapt to the presence of noise in human instructions or even locate and refine potentially inconsistent information?
% \item How to explicitly incorporate the natural of design diversity and develop techniques for diverse floor plan generation?
% \item How to extend the language-guided floor plan generation task to more practical but challenging scenarios, where users/clients may randomly forget to describe several room features or even the entire room?
% \squishend

\section{Conclusion}
In this paper, we initiate the research of a novel language-guided design generation task, with a specific focus on the floor plan domain as a start. We formulate it as \textit{language-guided floor plan generation} and introduce \textit{Tell2Design} (T2D), a large-scale dataset that features floor plans with natural language instructions in describing user preferences. We propose a Seq2Seq model as a strong baseline and compare it with several text-conditional image generation models. Experimental results demonstrate that the design generation task brings up several challenges and is not well-solved by existing text-conditional image generation techniques.
%our proposed method can generate reasonable floor plans following the given language instructions.  
Human evaluations assessing the degree of alignment between text and design, along with the human performance on the task, expose the challenge of understanding fuzzy and entangled information, and the nature of design diversity in our task.
%Human evaluations assessing the degree of alignment and analysis of human performance further exhibit the task features and expose the nature of design diversity in our task.
We hope this paper will serve as a foundation and propel future research on the task of the language-guided design generation.

\newpage
\section*{Limitations}
The proposed T2D dataset has several limitations, which could be addressed in future work. First, it only considers and collects language instructions for the floor plan domain. Future work could extend this language-guided design generation task to other design domains such as documents, mobile UIs, etc. Second, it is limited in the scope of languages where we only collect instructions written in English. Future work could assess the generalizability of the T2D dataset to other languages. Third, although generating floor plan designs from languages exhibit diversity, we do not consider improving generation diversity at this moment. Future works could consider building frameworks that specifically aim at design diversity. 

\section*{Ethics Statement}
In this section, we discuss the main ethical considerations of \textit{Tell2Design} (T2D): (1) Intellectual property protection. The floor plans of the T2D dataset are from the RPLAN \cite{wu2019data} dataset. Our dataset should be only used for research purposes. (2) Privacy. The floor plan data sources are publicly available datasets, where private data from users and floor plans have been removed. Language instructions are either generated artificially or collected from Amazon Mechanical Turk, a legitimate crowd-sourcing service, and do not contain any personal information. (3) Compensation. During the language instruction collection, the salary for annotating each floor plan is determined by the instruction quality and Mturk labor compensation standard.

\section*{Acknowledgements}
\textcolor{black}{We would like to thank the anonymous reviewers, our meta-reviewer, and senior area chairs for their constructive comments and support of our work. 
We also gratefully acknowledge the support of NVIDIA AI Technology Center (NVAITC) for our research.
This research/project is supported by the National Research Foundation Singapore and DSO National Laboratories under the AI Singapore Program (AISG Award No: AISG2-RP-2020-016).}

\bibliography{anthology,custom}
\bibliographystyle{acl_natbib}

\appendix

\section{Implementation Details}
\label{sec: appendix implementation details}
\paragraph{T2D Parameters}
In practice, we initialize all weights of our proposed baseline method from T5-base\footnote{\url{https://huggingface.co/t5-base}}. In training, we use Adam \cite{kingma2014adam} with $\beta_1=0.9, \beta_2=0.999, \epsilon=1e-08$ to update the model parameters. We fine-tune our model on $3$ RTX 8000 GPUs with batch size $12$ and learning rate $5e-4$ for $20$ epochs.

\paragraph{Baseline Implementation}
For the mentioned baselines, only Obj-GAN and CogView are open-sourced. Therefore, we adapt and implement the models from their official GitHub repositories\footnote{\url{https://github.com/jamesli1618/Obj-GAN}; \url{https://github.com/THUDM/CogView}}. However, as Imagen's source codes are not published, we implement it from the most starred GitHub repository\footnote{\url{https://github.com/lucidrains/imagen-pytorch}} (\ie, $5.9k$ stars until writing this paper) and adapt it to our T2D dataset. We use the process floor plan images in Graph2Plan \cite{hu2020graph2plan} for training. Although all baselines have provided pre-trained checkpoints for fine-tuning, our preliminary experiments indicate that training those baselines from scratch on the T2D dataset will obtain better performances. One most probable reason is the huge discrepancy between the data distributions of the baseline pre-training corpus and our T2D dataset. Those baseline checkpoints are mostly trained with real-life images with various objects and backgrounds. But our T2D dataset only focuses on the floor plan domain. 

Specifically, for Obj-GAN, we adopt and freeze the pre-trained text encoder, and train the rest of the networks (e.g., LSTMs) from scratch. For CogView, we freeze the pre-trained VQ-VAE and initialized the main backbone, decoder-only transformer, from GPT \cite{radford2019language,brown2020language}. During training, only the parameters of the transformer backbone will be updated. For Imagen, we import the T5-large model's encoder from \textit{Hugging Face} for text encoding and freeze all its parameters during training. The rest U-nets for diffusion will be updated according to the loss propagation.

\section{Dataset Analysis}
\label{sec: appendix dataset analysis}
\begin{figure}[t]
    \centering
    \includegraphics[scale=0.38]{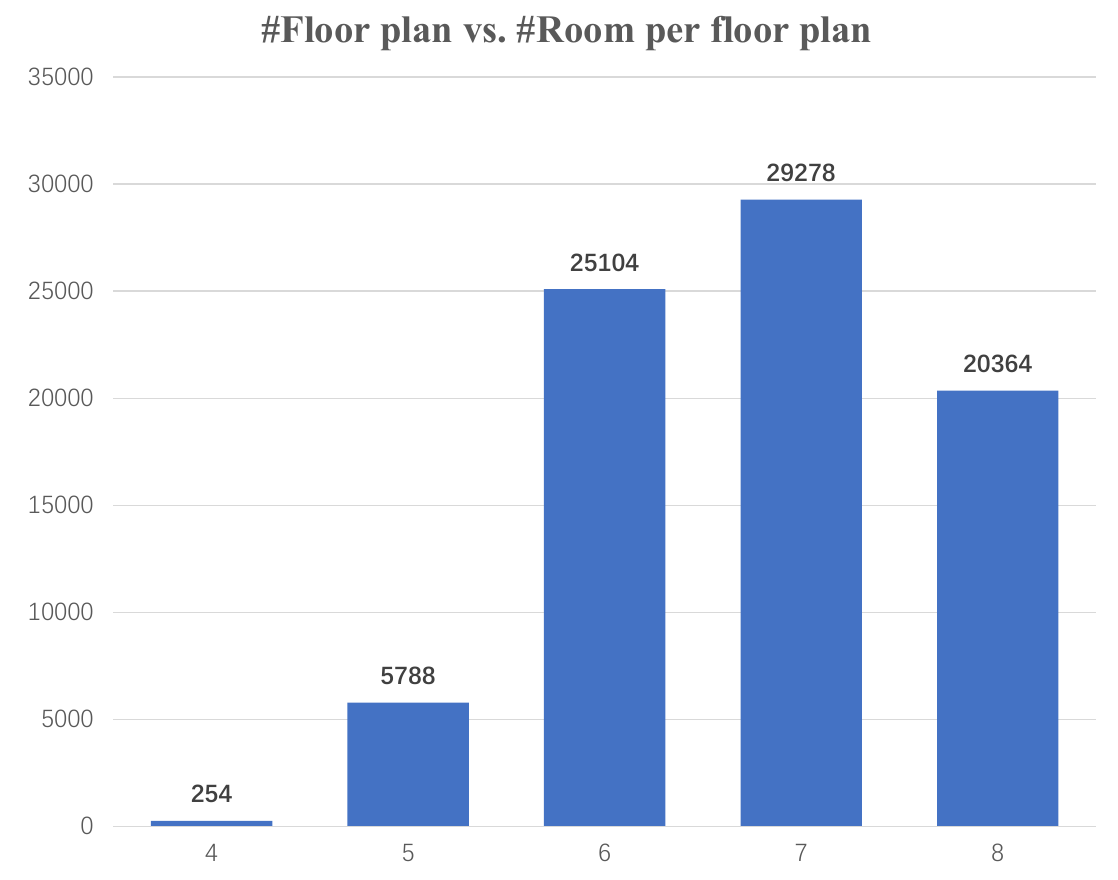}
    \caption{Number of floor plans vs. number of rooms per floor plan in the T2D dataset.} 
    \label{fig: room numbers}
\end{figure}

\paragraph{Floor Plan Statistics}
As shown in Tabel~\ref{tab: room occurrence} and Figure~\ref{fig: room numbers}, we present the statistics on the occurrence of each room type and the number of rooms per floor plan. There are 8 types of rooms in total. More than $92\%$ of floor plans include at least $6$ distinct rooms, and the most frequent room types are \textit{Common Room, Bathroom, Balcony, Living Room, Master Room, and Kitchen.}

\begin{table}[t]
\begin{center}
\centering
\small
\resizebox{0.65\columnwidth}{!}{%
\begin{tabular}{lr}
\toprule
\textbf{Room type}  & \#Floor plan \\ \midrule
CommonRoom & 100,847      \\
Bathroom   & 97,113       \\
Balcony    & 86,545       \\
LivingRoom & 80,788       \\
MasterRoom & 80,466       \\
Kitchen    & 77,768       \\
Storage    & 3,351        \\
DiningRoom & 1,312        \\ \bottomrule
\end{tabular}
}
\end{center}
\caption{Room type occurrences in the T2D dataset.}
\label{tab: room occurrence}
\end{table}

\paragraph{Dataset Comparison}
As shown in Table~\ref{tab: T2Dvslayout}, compared with other-related layout generation datasets, our T2D dataset is the first to have language annotations and aims to generate layout designs directly from languages. Since generating floor plan designs from language instructions can be naturally formulated into a text-conditional image generation problem, we compare our dataset with two benchmarking text-conditional image generation datasets in Table~\ref{tab: T2DvsCV}. We observe that our dataset is with a similar number of images with MSCOCO and Flickr30K but contains far longer text annotation for each image (\ie T2D has $256$ words on average describing each floor plan image). Moreover, as a set of language instructions for a floor plan results in a document-level text description, we compare our dataset with other document-level NLP datasets in Table~\ref{tab: T2DvsNLP}. We hope that our dataset can also propel the research on document-level language understanding. It is shown that our dataset has a comparable total number of samples and words with the largest DocRED\cite{yao2019docred}. More importantly, our “documents" are either human-annotated or artificially generated, instead of being crawled from the internet.

\begin{table}[t]
\begin{center}
\centering
\small
\resizebox{0.8\columnwidth}{!}{%
\begin{tabular}{lrr}
\toprule
\textbf{Dataset}   & \# Img.   & Avg. \# Words  \\ \midrule
MS COCO   & 82,783        & 11.3           \\
Flickr30K & 31,000        & 11.8           \\
T2D (ours) & 80,788        & 256.7          \\ \bottomrule
\end{tabular}
}
\end{center}
\caption{Comparisons between our T2D dataset and text-conditional image generation datasets.}
\label{tab: T2DvsCV}
\end{table}

\begin{table}[t]
\begin{center}
\centering
\small
\resizebox{1\columnwidth}{!}{%
    \begin{tabular}{lrrr}
    \toprule
    \textbf{Dataset}         & \# Doc. & \# Word & \# Sent. \\ \midrule
    SCIERC          & 500     & 60755   & 2217      \\
    BC5CDR          & 1,500   & 282k    & 11,089   \\
    DocRED (Human)  & 5,053   & 1,002k  & 40,276  \\
    DocRED (Distantly) & 101,873 & 21,368k & 828,115 \\
    T2D (Human)      & 5,051  & 1,011k     & 60.057         \\ 
    T2D (Artificial)   & 75,737  & 19,727k     & 1,776k         \\\bottomrule
    \end{tabular}
    }
\end{center}
\caption{Comparisons between our T2D dataset and document-level NLP datasets.}
\label{tab: T2DvsNLP}
\end{table}

\begin{table*}[t]
\begin{center}
\centering
\small
\resizebox{1\textwidth}{!}{%
    \begin{tabular}{lcccc}
    \toprule
    \textbf{Dataset}   & Domain               & Basic objects             & Object annotations        & Other annotations     \\ \midrule
    PubLayNet & scientific documents & \{text,title,figure,...\} & bounding boxes            & None                  \\
    RICO      & mobile UIs       & \{button,tool bar,...\}    & bounding boxes,interactions & animations,hierarchies \\
    SUB RGB-D & 3D indoor scenes & \{chair,table,pillow,...\} & bounding boxes              & 2D\&3D polygons        \\
    ICVT      & poster designs       & \{text,logo,...\}         & bounding boxes,substrates & background images     \\
    T2D (ours) & floor plans          & \{kitchen,balcony,...\}   & bounding boxes            & language instructions \\ \bottomrule
    \end{tabular}
    }
\end{center}
\caption{Comparisons between our T2D dataset and other layout generation datasets.}
\label{tab: T2Dvslayout}
\end{table*}

\section{Dataset Collection Details}
\label{sec: appendix dataset collection}
\begin{figure*}
    \centering
    \includegraphics[width=1.95\columnwidth]{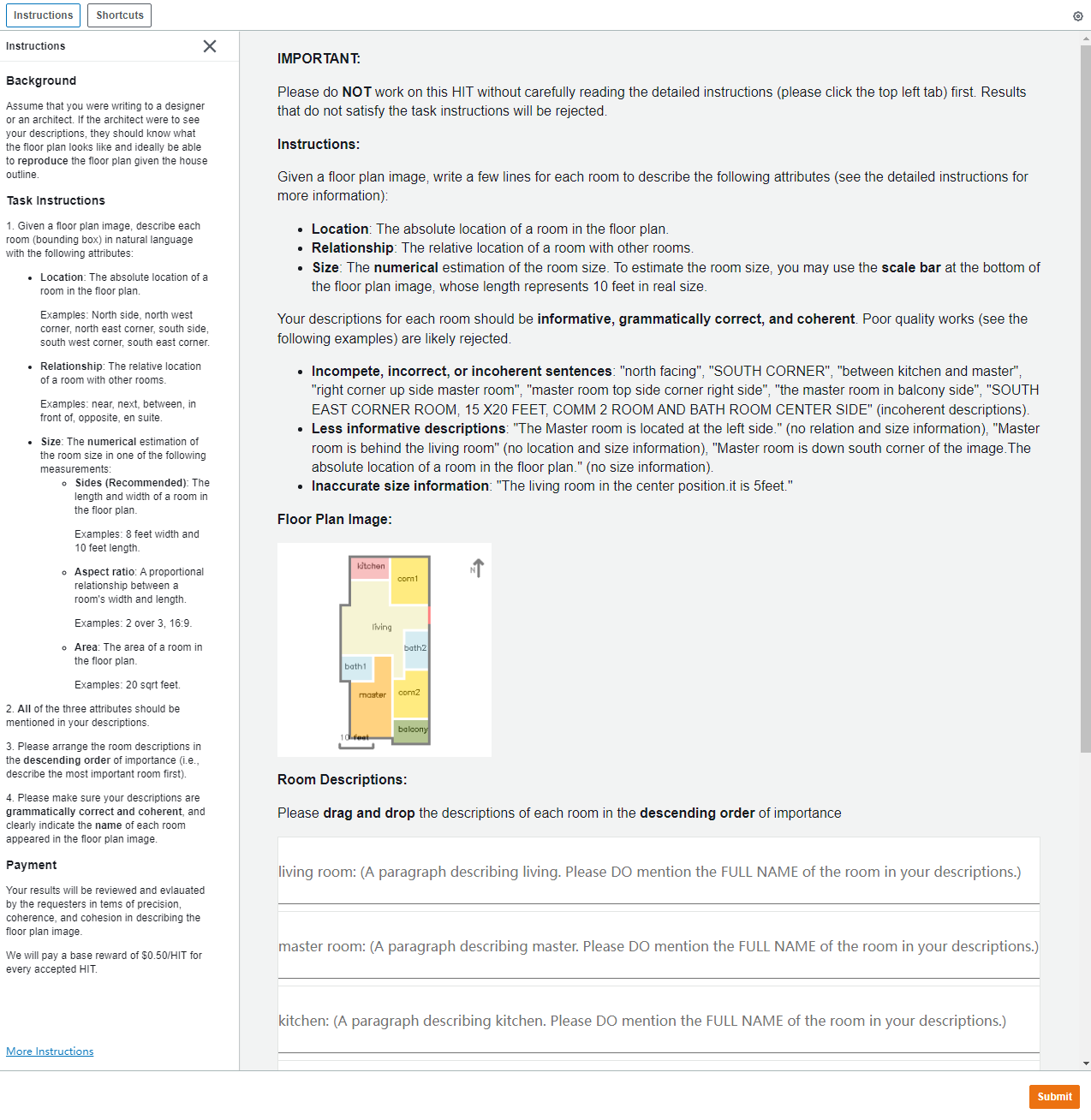}
    \caption{A screenshot of our HITs GUI.} 
    \label{fig: HITs}
\end{figure*}

\begin{figure*}
    \centering
    \includegraphics[width=1.95\columnwidth]{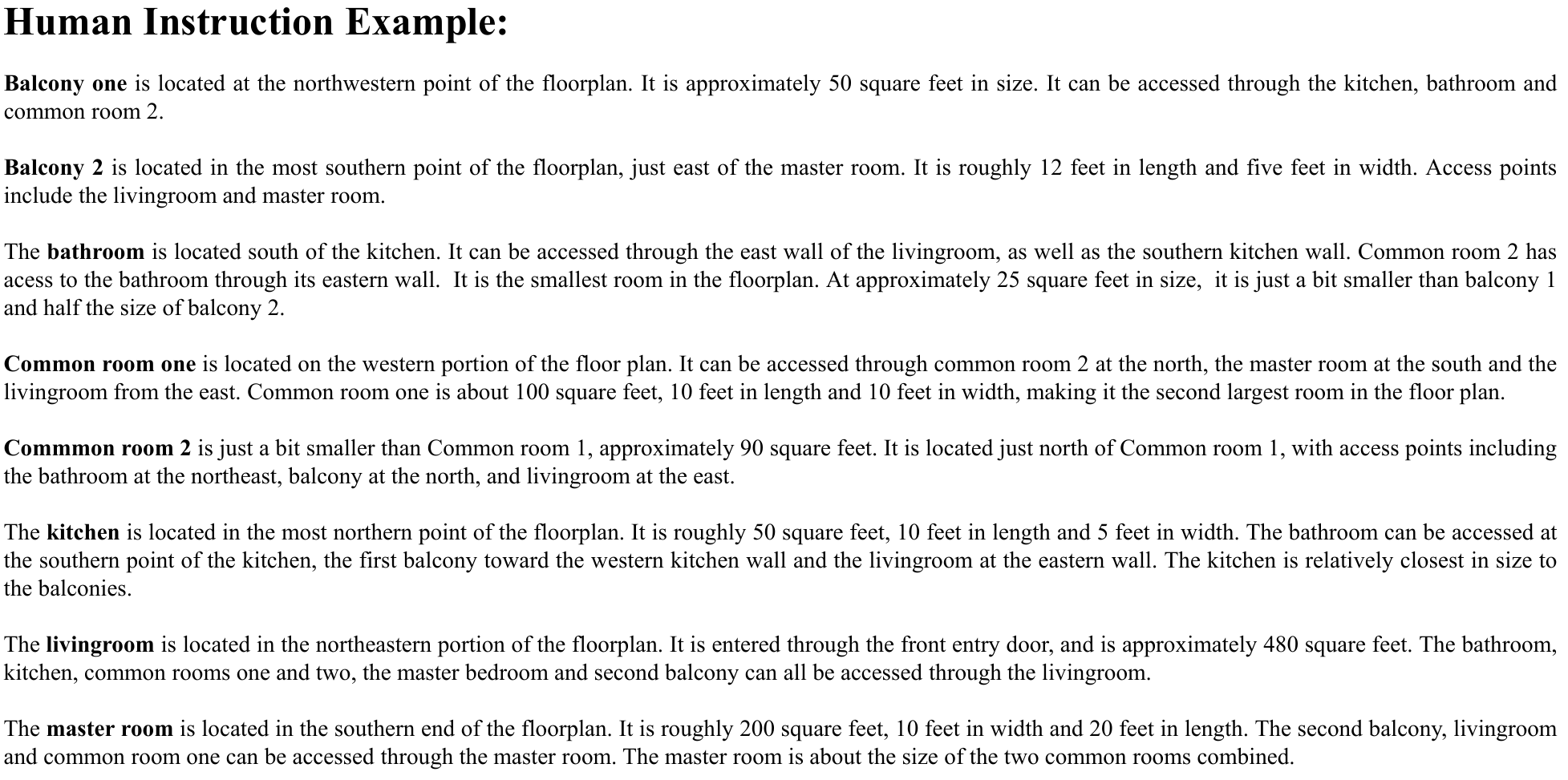}
    \caption{An example of human-written language instructions from the T2D dataset.} 
    \label{fig: human example}
\end{figure*}

\paragraph{Human instruction}
We employ Amazon Mechanical Turk (MTurk)\footnote{\url{https://www.mturk.com/}} to let annotators write language instructions for a given RGB 2D floor plan. Amazon considers this web service "artificial intelligence," and it is applied in various fields, including data annotation, survey participation, content moderation, and more. The global workforce (called "turkers" in the lingo) is invited for a small reward to work on "Human Intelligence Tasks" (HITs), created from an XML description of the task from business companies or individual sponsors (called "requesters"). 
HITs can display a wide variety of content (e.g., text and images) and provide many APIs, e.g., buttons, checkboxes, and input fields for free text.
In our case, turkers are required to fill the blank input fields in HITs with language instructions for each room, following our guidelines. 
% We encourage turkers to include (but not limited to) attributes such as room types, locations, sides, and relationships in their instructions.
% The definitions of these attributes are given as follows: The room type (e.g., bathroom and kitchen) specifies the functionality of a room. The room location specifies the global location of a room in the floor plan and can be described by phrases such as ``\texttt{north side}'' and ``\texttt{southeastern corner}''. The room sides specify the length and width of a room (e.g., ``\texttt{8 feet wide and 10 feet long}''). The room relationships specify the relative position of a room with other rooms and can be described by phrases such as ``\texttt{next to}'', ``\texttt{between}'', and ``\texttt{opposite}''. To mimic the real-world scenarios, we do not provide any bounding box information and ask crowdworkers to make rough estimations of the room size from the given floor plan image only. We also do not restrict the format of text descriptions or require all the above attributes to be mentioned, leading to unstructured and diverse instructions. Moreover, each annotated human instruction is manually assessed to ensure its quality. 
A screenshot of one of our HITs is displayed in Figure~\ref{fig: HITs}. We also show a full example of human instructions in Figure~\ref{fig: human example}.

\begin{figure*}
    \centering
    \includegraphics[scale=0.5]{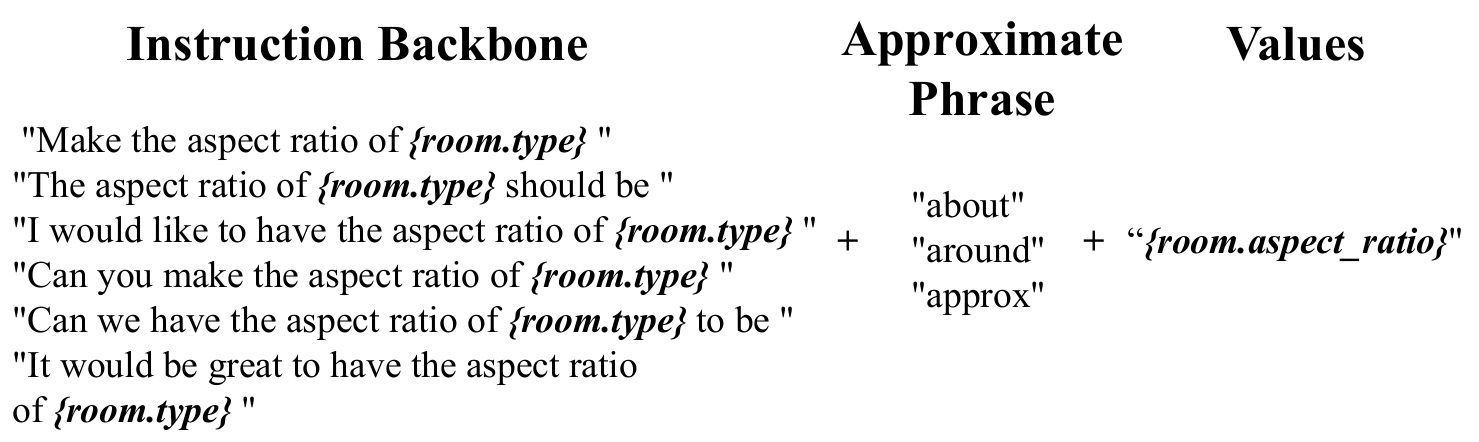}
    \caption{Illustration on generating artificial instructions describing room's aspect ratio.} 
    \label{fig: artificial template}
\end{figure*}

\begin{figure*}
    \centering
    \includegraphics[width=1.95\columnwidth]{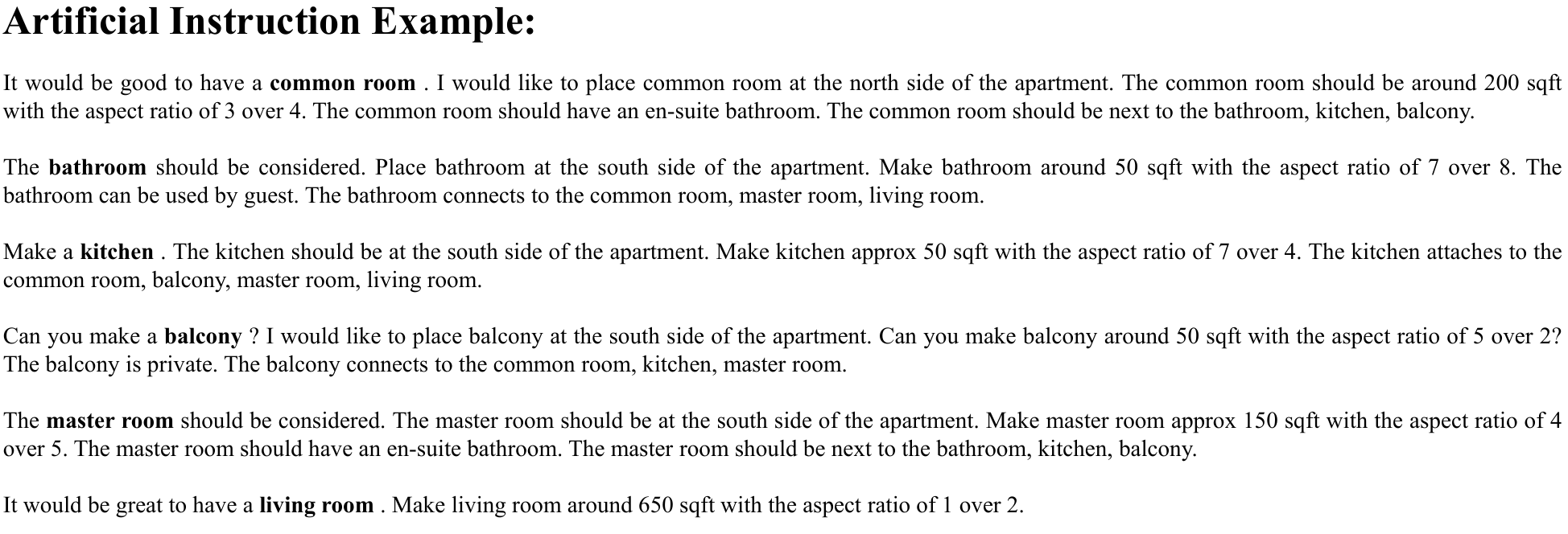}
    \caption{An example of artificially-generated language instructions from the T2D dataset.} 
    \label{fig: artificial example}
\end{figure*}

\paragraph{Artificial Instruction}
The artificial instructions in our T2D dataset are generated from scripts with several pre-defined templates.
\textcolor{black}{ We carefully select volunteers with natural language processing backgrounds for drafting templates.
% \textcolor{black}{Our volunteers for drafting templates were carefully selected based on their {\color{red}experience and proficiency in natural language processing}. }
Before participating in the annotation process, each annotator was required to undergo a qualification round consisting of a series of test annotations.}
We illustrate how we generate an instruction to describe one room's aspect ratio in Figure~\ref{fig: artificial template}. We also show a full example of artificial instructions in Figure~\ref{fig: artificial example}.
% put the final templates in Appendix.

\section{Baseline Generation Samples}
\label{sec: appendix baseline}
\begin{figure*}
    \centering
    \includegraphics[scale=0.5]{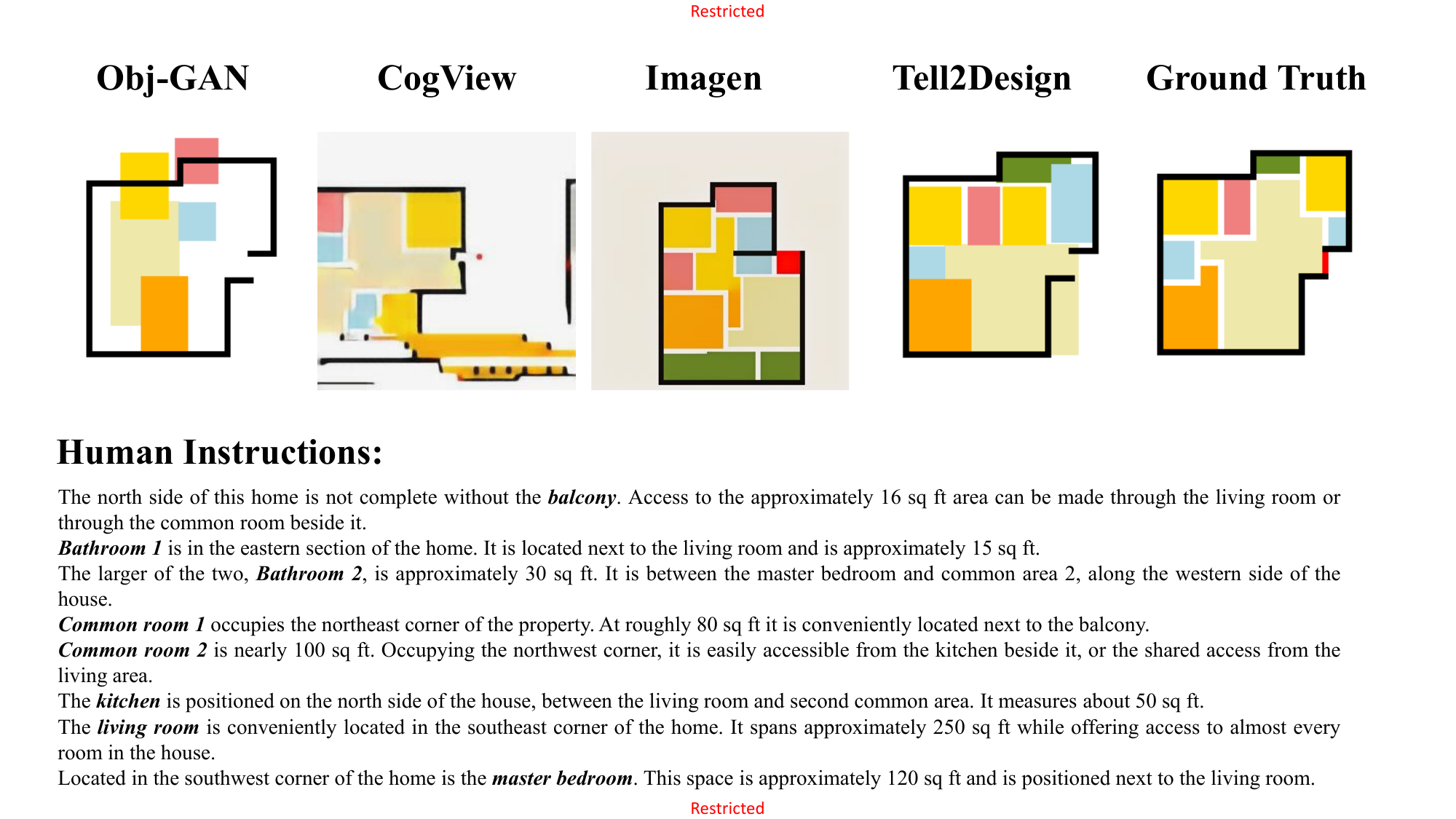}
    \caption{Generated samples from different baselines according to the same human-written language instructions.} 
    \label{fig: baseline}
\end{figure*}

To better understand and compare different baselines, we provide a case study of generated samples for the same language instructions from all baselines shown in Figure~\ref{fig: baseline}. 
Obj-GAN \cite{li2019object} has difficulties in capturing salient information from the given language instructions, resulting in generating rooms with incorrect attributes and relationships. One possible reason could be that it does not utilize any pre-trained large language model and thus suffers from understanding the given document-level instructions.
CogView \cite{ding2021cogview} instead auto-regressively generates the image tokens conditioned on all input instructions with a pre-trained GPT as the backbone. However, the image tokens sampled near the end of the generation show confusing information, resulting in an incomplete design. This is probably because presenting the whole floor plan design as a sequence of image tokens hinders the potential connections among different elements in the floor plan.
Imagen \cite{saharia2022photorealistic} exhibits its strong ability to generate realistic images in the target domain. However, it also fails to meet various design requirements specified in language instructions, indicating its limitation for design generation under multiple stricter constraints. 
%We can observe that Obj-GAN which does not utilize pre-trained large language models as its text encoder suffers from understanding the given language instructions, and thus result in 
\end{document}